\pdfoutput=1

\documentclass[11pt]{article}

\usepackage[final]{emnlp2021}

\usepackage{times}
\usepackage{latexsym}

\usepackage[T1]{fontenc}

\usepackage[utf8]{inputenc}

\usepackage{microtype}
\usepackage{graphicx}
\usepackage{multicol}
\usepackage{multirow}

\usepackage{amsmath}

\usepackage{array}

\title{One Agent To Rule Them All: Towards Multi-agent Conversational AI}

\author{
  Christopher Clarke$^1$\hspace{10pt} Joseph J. Peper$^1$\hspace{10pt} Karthik Krishnamurthy$^2$\hspace{10pt} Walter Talamonti$^2$ \\ \textbf{Kevin Leach$^3$\thanks{\hspace{1mm} Work was done while at University of Michigan}\hspace{10pt} Walter Lasecki$^*$\hspace{10pt} Yiping Kang$^1$\hspace{10pt} Lingjia Tang$^1$\hspace{10pt} Jason Mars$^1$}   \vspace{0.3cm}\\
    \text{$^1$University of Michigan, Ann Arbor, MI}\\
    \text{$^2$Ford Motor Company, Dearborn, MI}\\
    \text{$^3$Vanderbilt University, Nashville, TN}\\
    \text{\{csclarke, jpeper, ypkang, lingjia, profmars\}@umich.edu} \\
    \text{\{kkrish65, wtalamo1\}@ford.com},
    \text{kevin.leach@vanderbilt.edu, wslasecki@gmail.com}
}

\begin{document}
\maketitle
\begin{abstract}
The increasing volume of commercially available conversational agents (CAs) on the market has resulted in users being burdened with learning and adopting multiple agents to accomplish their tasks.
Though prior work has explored supporting a multitude of domains within the design of a single agent, the interaction experience suffers due to the large action space of desired capabilities.
To address these problems, we introduce a new task BBAI: \textbf{B}lack-\textbf{B}ox \textbf{A}gent \textbf{I}ntegration, focusing on combining the capabilities of multiple black-box CAs at scale.
We explore two techniques: \emph{question agent pairing} and \emph{question response pairing} aimed at resolving this task.
Leveraging these techniques, we design One For All (OFA), a scalable system that provides a unified interface to interact with multiple CAs.
Additionally, we introduce MARS: \textbf{M}ulti \textbf{A}gent \textbf{R}esponse \textbf{S}election, a new encoder model for question response pairing that jointly encodes user question and agent response pairs.
We demonstrate that OFA is able to automatically and accurately integrate an ensemble of commercially available CAs spanning disparate domains.
Specifically, using the MARS encoder we achieve the highest accuracy on our BBAI task, outperforming strong baselines.
\end{abstract}

\section{Introduction}

Influenced by the popularity of intelligent conversational agents (CAs), such as Apple Siri and Amazon Alexa, the conversational AI market is growing at an increasingly rapid pace and is projected to reach a valuation of US \$13.9 billion by 2025~\cite{market_and_markets_2020}.
These CAs have already begun to show great promise when deployed in domain-specific areas such as driver assistance~\cite{Lin2018Adasa}, home automation~\cite{Luria2017}, and food ordering~\cite{foodbot} with platforms such as Pandora and Facebook today hosting more than 300,000 of these agents~\cite{Chaves:2018:SMC:3173574.3173765,nealon_2018}.

\begin{figure}
  \centering
  \includegraphics[width=\columnwidth]{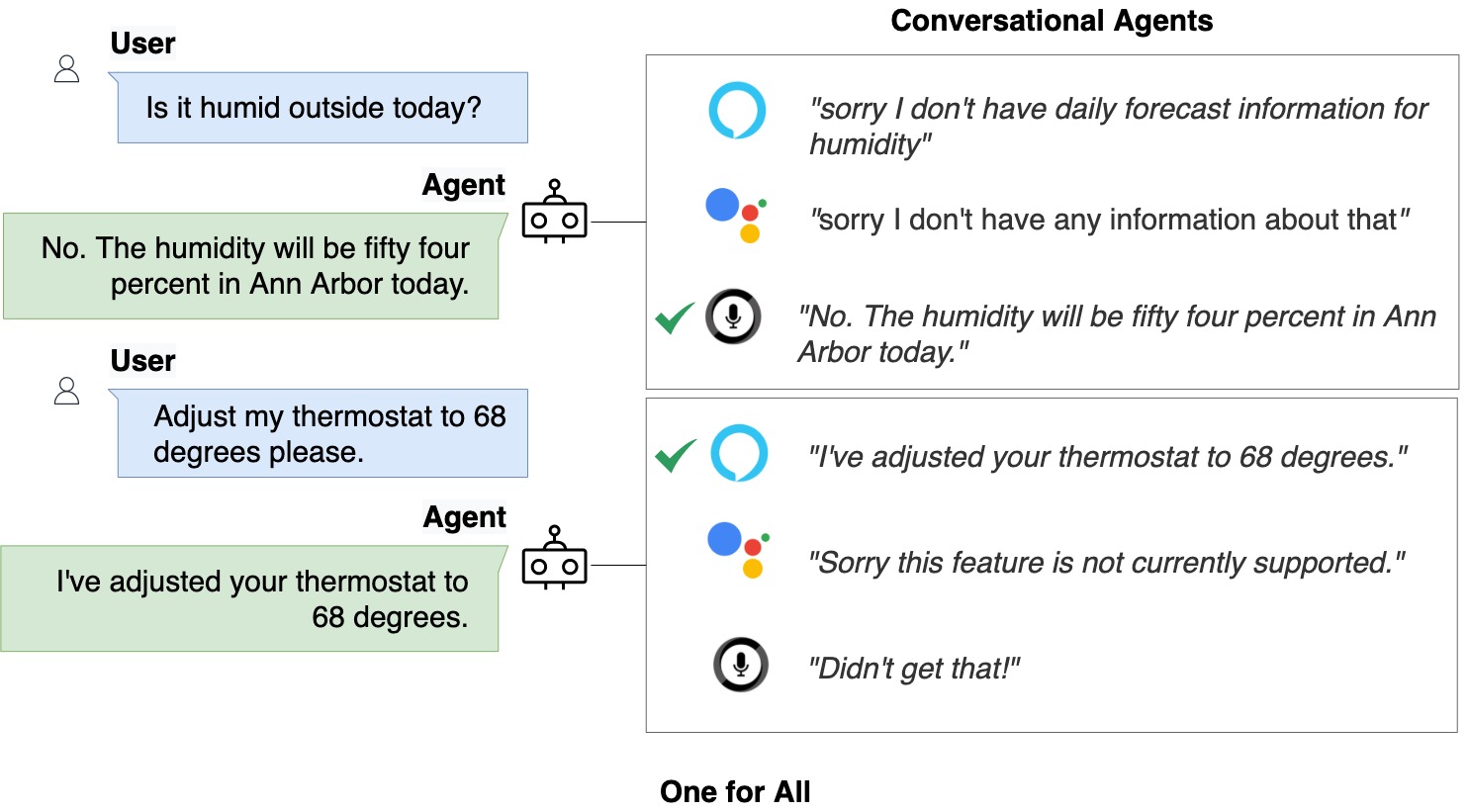}
  \vspace{-1.8pc}
  \caption{An example interaction using One For All which integrates multiple production black-box agents into a unified experience.  }
  ~\label{fig:ofa}
  \vspace{-2.8pc}
\end{figure}

Most CAs are designed to be specialized in a single or set of specific domains.
As such, users are required to interact with multiple agents in order to complete their tasks and answer their queries as shown in figure \ref{fig:ofa}. For example, a user may use an agent such as Amazon Alexa for online shopping but engage with Google Assistant for daily news updates. Additionally, a given agent may be more proficient at a specific domain over another i.e A finance CA is better suited to answer finance questions. As a result, users are taxed with the burden of learning and adopting multiple agents leading to an increase in the cognitive load of interacting with agents, further discouraging the proliferation of their usage \cite{dubiel2020, novick2018, choice2019}. This is escalated further as the number of conversational agents deployed into the market continues to increase.
Therefore, the need arises for unifying multiple independent CAs through one conversational interface. 
This need has manifested in the commercial conversational AI industry with initiatives such as the Amazon Voice Interoperability Initiative~\cite{amazon.com} which aims to create voice-enabled products that contain multiple, distinct, interoperable intelligent assistants on a single device.
However, this interaction is still manual, requiring the user to orchestrate which agent is initiated. In addition, while it is possible to have distinct agents in a single device, users prefer interacting with a single agent over multiple~\cite{Chaves:2018:SMC:3173574.3173765}.

Prior work has explored in part combining the strengths of multiple agents in one system but they rely on direct access to the design and implementation details of the to-be-integrated agents. \citet{Subramaniam:2018:CCM:3237383.3237472} and \citet{alanav2} direct incoming user questions to a specific agent based on the candidate agents' internal knowledge graph and NLU architectures, respectively. However, in practice, the majority of the publicly available CAs are "black boxes" where their inner-workings contain highly-protected IP that is not accessible to the public. Additionally, \citet{alanav2} facilitates their bot selection with a manual heuristic preference order that requires intimate knowledge of the agents to construct, and additional effort to maintain, thus not scaling well for the adaption of existing agents and introduction of new agents. Therefore, the task of integrating multiple production black-box CAs with a unified interface remains an open problem.

In order to explore this problem, we introduce the task BBAI: \textbf{B}lack-\textbf{B}ox \textbf{A}gent \textbf{I}ntegration that focuses on integrating multiple black-boxes CAs. We propose two techniques to tackle black-box multi-agent integration: (1) Question agent pairing and (2) Question response pairing. Intuitively, these two approaches can be viewed as a query-to-agent classification problem in contrast to that of a response selection problem. This formulation allows us to facilitate multi-agent integration whilst operating within the black-box constraints of the agents.
Using these techniques we develop \emph{One For All}, a novel conversational system that accurately and automatically unifies a set of black-box CAs spanning disparate domains. Additionally, we introduce MARS: \textbf{M}ulti \textbf{A}gent \textbf{R}esponse \textbf{S}election, a new encoder model for question response pairing that jointly encodes user question and agent response pairs.
We evaluate these techniques on a suite of 19 publicly available agents consisting of Amazon Alexa\footnote{\url{https://developer.amazon.com/en-US/alexa}}, Google Assistant\footnote{\url{https://assistant.google.com/}}, SoundHound Houndify\footnote{\url{https://www.houndify.com/}}, Ford Adasa~\cite{Lin2018Adasa} and many more.

Specifically, this paper makes the following contributions:

\begin{itemize}
    \item Formulation of the BBAI task that focuses on the challenge of integrating disparate black-box conversational agents into one experience. We construct a new dataset for this task, comprising of examples from a suite of 19 commercially deployed conversational agents. We publish our model and datasets. \footnote{\url{https://github.com/ChrisIsKing/black-box-multi-agent-integation}}
    
    \item We design \emph{One For All}, a novel conversational system that accurately and automatically unifies a set of black-box CAs and introduce the MARS encoder model that outperforms strong state-of-art classification and ranking model baselines on our BBAI task.
    
    \item We conduct a thorough evaluation of various agent integration approaches showing that our MARS encoder outperforms strong baselines. We show that by facilitating the integration of multiple agents we can alleviate the need for users to adopt multiple agents whilst facilitating the improvement and growth of agents over time.
\end{itemize}

\section{BBAI: Black-Box Agent Integration Task Formulation} \label{sec_task}
 Building a unified interface for production agents spanning different domains presents several key challenges. 
 First, most commercially available CAs are black-boxes, providing little to no information on their inner workings. Any approach for agent integration must operate without relying on the internals of any given agent.
 Second, these conversational agents are constantly improved upon and expanded with new capabilities. The agent integration approaches need to be flexible and adaptive to these changes with relative ease. Given these constraints we assume the existence of the following information sources for the agent integration task:
 \begin{enumerate}
     \item User query/utterance: The question that the user asks the agent.
     
     \item Agent skill representation: A textual representation that denotes what each agent is capable of. This can be in the form of example queries or a description of that agent.
     
     \item Agent response: Each agent's response to the query asked.
 \end{enumerate}

\begin{figure}
  \centering
  \includegraphics[width=\columnwidth]{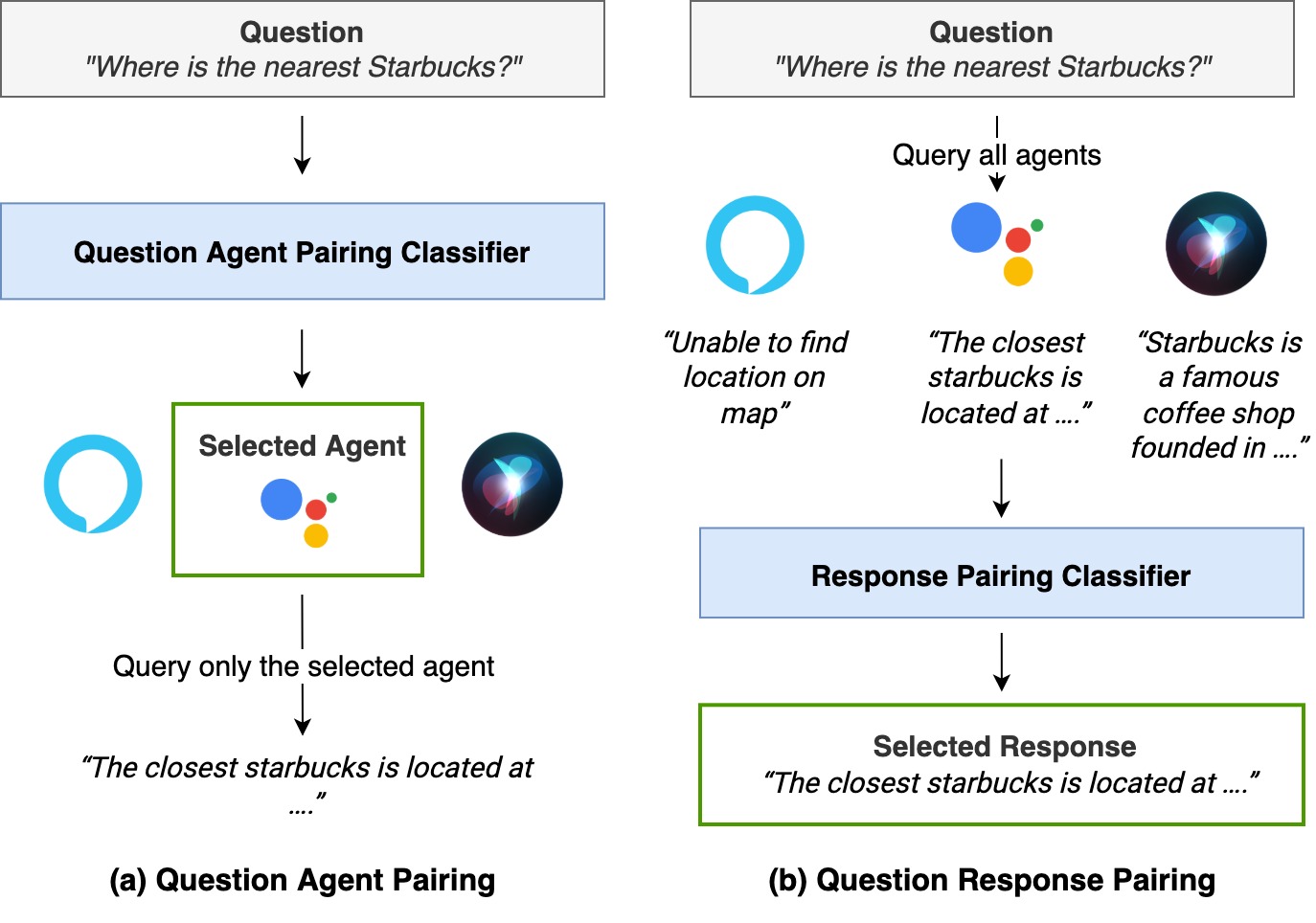}
  \caption{Overview of our proposed black-box agent integration techniques. In QA Pairing, the goal is to select the correct agent using information about the agent's capabilities. In QR Pairing, the goal is to select the correct agent response.}
  ~\label{fig:task}
  \vspace{-1.8pc}
\end{figure}

Using this information we formulate the task of agent integration as given a query $Q$, a set of agents $A = \{a_{1}, a_{2}, \ldots, a_{n}\}$ and a set of agent responses $R = \{r_{1}, r_{2}, ..., r_{n}\}$ to query $Q$, determine the question-agent-response pair $(Q, A_{i}, R_{i})$ that resolves the query $Q$. Further, given the information available, we can taxonomize our approach into  two techniques: (1) Question agent pairing where we preemptively select the agent for the query and (2) Question response pairing where we evaluate the set of returned responses as depicted in Figure \ref{fig:task}.

\subsection{Question Agent Pairing} \label{sec_qap}
As shown in Figure~\ref{fig:task}, the goal of question agent pairing is, given a query $Q$ and a set of agents $A = \{a_{1}, a_{2}, \ldots, a_{n}\}$, determine the question-agent pair $(Q, A_{i})$ that resolves the query $Q$. At its core, this can be viewed as a classification problem where the model learns the respective capabilities of each independent agent in order to predict which agent to use for a given question.

\subsection{Question Response Pairing}
As shown in Figure~\ref{fig:task}, the goal of question response pairing is, given a query $Q$ and a set of agent responses $R = \{r_{1}, r_{2}, ..., r_{n}\}$, determine the question-response pair $(Q, R_{i})$ such that $R_i$ resolves the query $Q$.

 \section{The One For All System}

 In this section, we present the design of One For All (OFA), a scalable system that integrates multiple black-box CAs with a unified interface. We explain the various approaches implemented in One For All, detailing their inputs, outputs and training methodology.
 
  \begin{figure}
  \centering
  \includegraphics[clip, trim=2cm .5cm 0cm 0cm, width=\columnwidth]{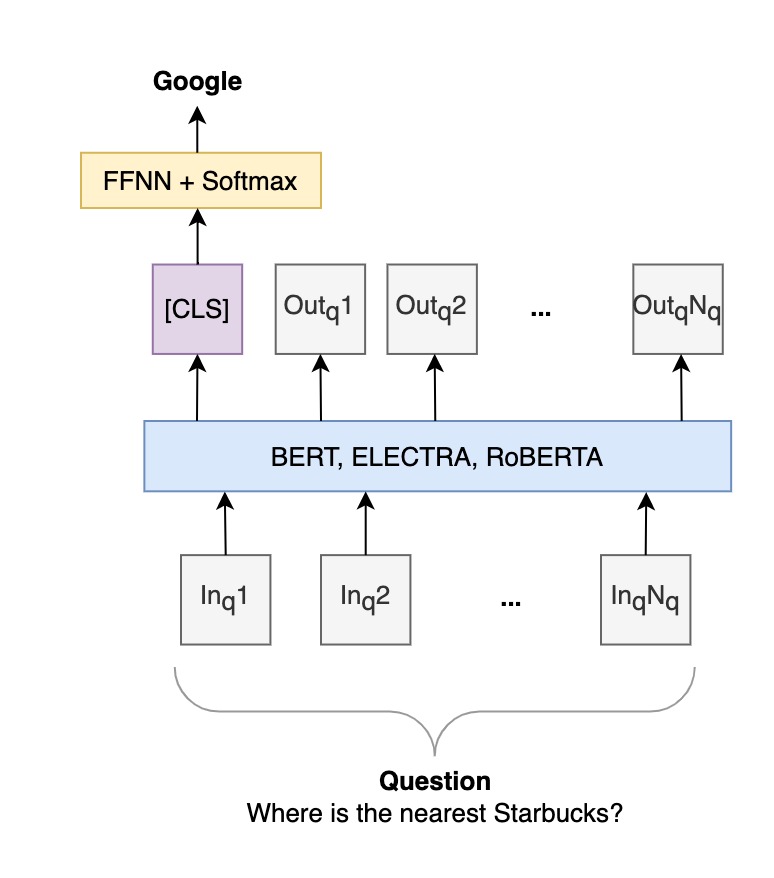}
  \caption{The transformer-based classification models in the OFA system. The models are trained on question agent pairs and tasked to predict an agent to route the given query to.}
  ~\label{fig:classifier}
  \vspace{-2pc}
\end{figure}
 
\begin{figure*}
  \centering
  \includegraphics[width=1\textwidth]{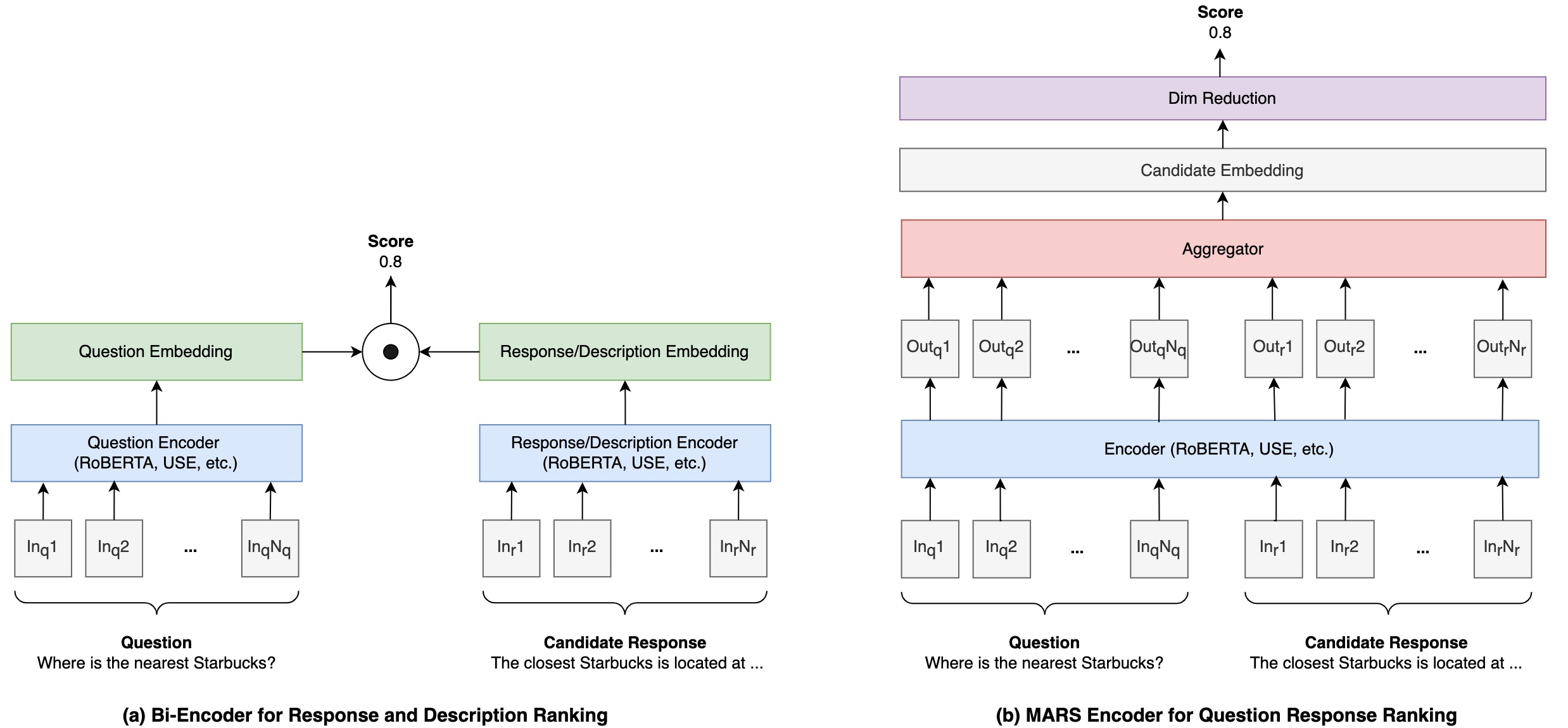}
  \caption{Overview of OFA approaches. (a) Bi-Encoder which is used for both QA and QR pairing encodes the question and candidate response/description separately and computes a ranking score via a dot product calculation. (b) Our MARS encoder jointly encodes the question and response into a single transformer and performs self-attention between the question and candidate response. To score a response we reduce the candidate embedding from a vector to a scalar score between 0...1  \cite{Humeau2020Poly-encoders:}.}~\label{fig:system}
  \vspace{-1.8pc}
\end{figure*}
 
 \subsection{Question Agent Pairing}
 In order to predict the best agent for a given query, knowledge of each agent's individual skill-set is required. However, as described in the task formulation in Section \ref{sec_task}, the internal details of the agents are unavailable. Everyday users of these agents have no insight into the internal specifics of these agents. However, they are able to use these agents to accomplish tasks by building a mental model of each agents' respective capabilities through usage over time. 
 We draw inspiration from this to determine the information we can use to represent an agent's skills without access to its internals.
 
 \subsubsection{Agent Skills Representation}
Following the learning patterns described above, we model an agent's skill-set in two distinct ways: 

\textbf{(1) Query examples:} Similar to building knowledge overtime via agent interaction, an agents' query examples allows the model to learn what type of queries each agent is capable of resolving. For example, questions such as \textit{``Where is the nearest gas station?"} and \textit{"Direct me to Starbucks please"} will be amongst the query examples for a \textit{``Directions"} agent.

\textbf{(2) Agent descriptions:}. These are textual summaries of an agent's capabilities. For example, a bank releases a new CA for its customers to use instead of having to visit the bank. Accompanied with this agent will be a semi-formal description of what this agent is capable of doing. This information is often publicly available in the agent's marketing materials.

Using these query examples and agent descriptions, we explore approaches for determining the agent best to resolve a given query. 
We describe in more detail the dataset collection process in Section~\ref{sec:data}.

\paragraph{Question agent pairing using query examples}
QA pairing using query examples seeks to explore how best we can facilitate agent orchestration in a data constrained environment where only a few examples of the questions the agents can answer are present. This is similar to the use of text examples for the training of an intent classifier but at the agent level instead. Therefore, we treat this as a multi-label classification problem where a given query $Q$ is mapped to a set of agents $A$. e.g Q: \emph{`locate me some good places in Kentucky that serve sushi`} maps to the set of agents $A$: [“Alexa”, "Google"] indicating that this query can be correctly answered by the agents Alexa and Google. Specifically, as shown in Figure \ref{fig:classifier}, we build a multi-label classifier on top of state-of-the-art transformer models, BERT \cite{devlin-etal-2019-bert}, RoBerta \cite{liu2019roberta} and Electra \cite{clark2020electra} to predict an agent $A$ given a query $Q$.



\paragraph{Question agent pairing using agent descriptions}
While query examples are useful for understanding the capabilities of a given agent, they may not be readily available. When a new agent is introduced, users are unsure of the exact questions this agent can answer but they would typically have access to an explanation of its capabilities. As an alternative, we explore the use of such a description of the agents.
For this task, we assume a textual description of an agent's capabilities, e.g. "Our productivity bot helps you stay productive and organized. From sleep timers and alarms to reminders, calendar management, and email ....".

In order to map a given query $Q$ to an agent $A$ described by description $D_{i}$, we treat this as a semantic similarity task. The intuition behind this is that for a given query $Q$ the agent that is capable of answering a given question is likely to feature an agent description semantically similar to the question. We explore a suite of pre-trained and fine-tuned language models focusing on ranking the relevance of given description $D_{i}$ to a query $Q$. Additionally, given the length of descriptions and the range of capabilities that may be described within a single description, we split the full description at the sentence level and use each sentence to represent a single skill $S_{i}$ belonging to agent $A$. With this variation, the question-description similarity score is calculated as the $\max_i{SemSim(Q, S_{i})}$.

For our BBAI task we consider the following state-of-art semantic retrieval-based approaches whose utility map well to our problem domain:

\paragraph{BM25} This classic method measures keyword similarity and uses it to estimate the relevance of documents to a given search query \cite{10.1561/1500000019}. We encode the collection of agent descriptions and return the agent whose description is most relevant to the given query.

\paragraph{Universal Sentence Encoder} ~\cite{cer-etal-2018-universal} A sentence encoding model for encoding sentences into high dimensional vectors. We use the transformer model\footnote{\url{https://tfhub.dev/google/universal-sentence-encoder/4}} for our experiment. As shown in part (a) of Figure \ref{fig:system}, we encode the user query and the agent description and compute the dot product as a ranking score.

\paragraph{Roberta + STS} \cite{reimers-2019-sentence-bert} We fine-tune Roberta-base on the STS benchmark dataset and use this model to encode our agent descriptions and user query. We compute the cosine similarity between the two vectors to compute a ranking score for each description as shown in Figure \ref{fig:system}.

\subsection{Question Response Pairing}
Contrary to question agent pairing which selects the agent beforehand, question response pairing assumes that we provide each agent in the ensemble the opportunity to respond to the query $Q$ and focus on selecting the best response from the set of returned responses. As such, we treat this as a response ranking problem of determining which question-response pair $(Q, R_{i})$ best answers the query $Q$. Prior work has shown strong performance on sentence pairing tasks such as this through the use of sentence encoders and language model fine-tuning \cite{henderson2019convert, Humeau2020Poly-encoders:, reimers-2019-sentence-bert}. We explore the use of these architectures in the domain of response selection with the goal of learning representations for correct question answering from diverse conversational agents.


\paragraph{BM25} Similar to our use of BM25 for question agent pairing we use it to rank each of our question response pairs.

\paragraph{USE and USE QA} \cite{yang2019multilingual} We apply the USE model from our agent pairing task to rank agent responses. In addition, we consider USE QA, an extended version of the USE architecture specifically designed for question-answer retrieval applications. We use the Bi-Encoder architecture as shown in Figure \ref{fig:system} (a).

\paragraph{Roberta + STS} We fine-tune Roberta-base on the STS benchmark dataset and use it to encode our question response pairs using the bi-encoder architecture in figure \ref{fig:system}.

\paragraph{MARS encoder} Pre-existing sentence pairing scoring models are tuned to score sentence pairs deemed semantically similar. However, in the case of conversational systems, an agent's response can be semantically similar but still incorrect. e.g Q: "What is the weather in Santa Clara today?", R: "Weather information is currently unavailable". These two sentences are semantically similar but the response does not resolve the query. In the MARS encoder, we focus on learning representation beyond similarity by also incorporating the correctness of agent responses. Using the cross-encoder architecture \cite{Humeau2020Poly-encoders:, reimers-2019-sentence-bert} shown in part (b) of Figure \ref{fig:system}, we train a question response pair scoring model for the task of ranking responses to a given query $Q$ generated by conversational agents. We concatenate both the input question and response performing full self-attention on the entire input sequence. By passing both the question and agent response through a single transformer, the agent response is able to attend to the user query and produce a more input sensitive representation of the question response embedding. Using the generated question response embedding vector we then convert it to a scalar score $S(Q, R_{i})$ between 0..1 via a linear layer. Our training objective is to minimize the Cross-Entropy loss between the correct agent responses and the negative agent responses to the query $Q$.



\begin{table*}[]
\small
\centering
\resizebox{14cm}{!}{
\begin{tabular}{|c|>{\raggedright}p{3cm}|>{\raggedright}p{3cm}|>{\raggedright}p{3cm}|p{3cm}|}
\hline
\multirow{2}{*}{\textbf{Question}} & \multicolumn{4}{c|}{\textbf{Agent Response}} \\ \cline{2-5}
 & \multicolumn{1}{c|}{\textbf{Alexa}} & \multicolumn{1}{c|}{\textbf{Google}} & \multicolumn{1}{c|}{\textbf{Houndify}} & \multicolumn{1}{c|}{\textbf{Adasa}}   \\ \hline
\multicolumn{1}{|p{3cm}|}{At how many miles will I run out of gas} & "here's something I found on the web according to freakonomics.com previously when cars got 8 to 12 miles ...." & "on the website post Dash gazette.com they say some popular car models can make it between 30 and 50 miles ....", & Didn't get that! & {\color[HTML]{008000}"With your current fuel economy of 28 MPG, you should be able to cover about 532 miles with the fuel you have."} \\ \hline

\multicolumn{1}{|p{3cm}|}{Is it gonna be warm Friday in Alhambra?} & "here's something I found on the web according to Wikipedia. Org Cobra is one of the 100 selected cities in India which will be developed ...." & {\color[HTML]{008000}"No, it won't be hot Friday in Alhambra, California. Expect a high of 21 and a low of 6."}, & {\color[HTML]{008000}"There will be a high of seventy degrees in Alhambra on Friday November twenty-seventh."} & "Out of scope!" \\ \hline
\end{tabular}}
\caption{Sample question agent responses from the One For All dataset. Responses highlighted in {\color[HTML]{008000}green} represent agent responses voted as correct by crowd workers.  }
\label{tab:data}
\vspace{-.5cm}
\end{table*}


\section{Dataset Construction} \label{sec:data}

For the task of BBAI, we construct a new dataset focusing on making it representative of real-world conversational agents at scale and covering a broad range of domains. 

Using Amazon Mechanical Turk and scenario/paraphrasing-based prompts~\cite{Kang:2018, larson}, we crowdsourced utterances across a range of agent skills/capabilities. These skills were extracted from public information sources describing each of the agents, in addition to observing their capabilities. Our dataset is comprised of utterances across 37 broad domain categories. These include domains such as \textit{Weather, Flight Information, Directions, Automobile}, etc. Crowd workers were paid \$0.12 for 5 utterances. These submitted utterances were then vetted by hand to ensure quality. Using the curated utterances, we then generated question responses by querying each agent to gather its response to the utterance. 

In order to generate ground truth samples on which of the question-response pairs $(Q, R_{i})$ correctly resolves the query $Q$ we launched a crowdsourcing task asking workers to indicate the candidate responses that best answer the question shown. Five workers were assigned to each response selection task and majority voting (>2) was used to label the gold responses. As such for each query $Q$ and the set of responses $R$ we were able to gather the necessary question-agent pairs $(Q, A_{i})$ and question-response pairs $(Q, R_{i})$ needed to evaluate our approaches.

\begin{table*}[]
\small
\centering
\resizebox{12cm}{!}{%
\begin{tabular}{|clc|rrrr|}
\hline
\multicolumn{3}{|l|}{} & \multicolumn{4}{c|}{\textbf{Agent Breakdown}} \\ \hline
\multicolumn{1}{|l}{} & \textbf{Method} & \multicolumn{1}{l|}{\textbf{Accuracy (n=4)}} & \multicolumn{1}{c}{\textbf{Alexa}} & \multicolumn{1}{c}{\textbf{Google}} & \multicolumn{1}{c}{\textbf{Houndify}} & \multicolumn{1}{c|}{\textbf{Adasa}} \\ \hline
\multirow{3}{*}{\begin{tabular}[c]{@{}c@{}}Question Agent Pairing \\ (QA Labels)\end{tabular}} & Bert & 68.31 & 37.98 & \textbf{40.93} & 18.49 & 2.6 \\
 & Electra & 67.86 & 35.28 & \textbf{42.01} & 20.11 & 2.6 \\
 & Roberta & \textbf{69.03} & 34.92 & \textbf{41.56} & 20.65 & 2.87 \\ \hline
\multirow{3}{*}{\begin{tabular}[c]{@{}c@{}}Question Agent Pairing \\ (Descriptions)\end{tabular}} & BM25 & 27.91 & 13.91 & 10.95 & 17.33 & \textbf{57.81} \\
 & USE & \textbf{47.84} & 13.20 & 28.82 & \textbf{52.42} & 5.56 \\
 & Roberta+STS & 39.40 & 18.94 & 22.35 & \textbf{51.35} & 7.36 \\ \hline
\multirow{5}{*}{Response Selection} & BM25 & 51.07 & 28.64 & 24.69 & 14.81 & \textbf{31.86} \\
 & USE & 72.89 & \textbf{34.20} & 27.65 & 22.98 & 15.17 \\
 & USE QA & 75.49 & \textbf{41.65} & 36.45 & 17.95 & 3.95 \\
 & Roberta+STS & 69.83 & 18.94 & 22.35 & \textbf{51.35} & 7.36 \\
 & MARS & \textbf{79.70} & 37.34 & \textbf{43.9} & 15.71 & 3.05 \\ \hline
\multirow{4}{*}{Individual Agents} & Alexa & 49.37 & \multicolumn{1}{c}{\textbf{-}} & \multicolumn{1}{c}{-} & \multicolumn{1}{c}{-} & \multicolumn{1}{c|}{-} \\
 & Google & \textbf{51.79} & \multicolumn{1}{c}{-} & \multicolumn{1}{c}{-} & \multicolumn{1}{c}{-} & \multicolumn{1}{c|}{-} \\
 & Houndify & 34.82 & \multicolumn{1}{c}{-} & \multicolumn{1}{c}{-} & \multicolumn{1}{c}{-} & \multicolumn{1}{c|}{-} \\
 & Adasa & 4.12 & \multicolumn{1}{c}{-} & \multicolumn{1}{c}{-} & \multicolumn{1}{c}{-} & \multicolumn{1}{c|}{-} \\ \hline
\end{tabular}
}
\caption{Performance breakdown of QA and QR approaches on our BBAI task when using our 4 largest agents Alexa, Google, Houndify and Adasa. \textbf{Note:} n = number of agents.}
\label{tab:results-4}
\end{table*}

\paragraph{Agent Descriptions}
We gather our agent descriptions by scraping the contents of each of the agent's public product pages and their built-in feature documentation web pages. We then manually clean, reformat and merge this data into a single document per agent. For our experiment, we focus only on extracting descriptions related to the built-in features of our agents.

Overall our dataset contains 5550 utterances with 19 question-response pairs per question (one from each of the 19 agents), 105,450 in total. The utterances are split into 3700 utterances (100 per domain) for the training set and 1850 (50 per domain) for the test set. The train and test sets respectively contain 2399 and 1186 utterances with at least one positive question-response pair. In the remaining examples, none of the agents were able to achieve annotator agreement (>= 3). A sample dataset example is shown in table \ref{tab:data} with responses from 4 of the 19 agents.

\begin{table}[]
\small
\centering
\resizebox{\columnwidth}{!}{%
\begin{tabular}{|clr|c|}
\hline
\multicolumn{2}{|c}{\textbf{Method}} & \multicolumn{1}{l|}{\textbf{Accuracy (n=19)}} & \textbf{Agents} \\ \hline
\multirow{3}{*}{\begin{tabular}[c]{@{}c@{}}Question Agent Pairing\\  (QA Labels)\end{tabular}} & Bert & 59.10 & \multirow{15}{*}{\begin{tabular}[c]{@{}c@{}}Alexa, Google\\ Houndify, Adasa\\ Recipe agent\\ Dictionary agent\\ Task Manager\\ Hotel agent, Stock agent\\ Math agent, Sports agent\\ Wikipedia agent\\ Mobile Account agent\\ Banking agent\\ Coffee shop agent\\ Event Search agent\\ Jokes agent\\ Reminders agent\\ Covid-19 agent\end{tabular}} \\
 & Electra & 52.86 &  \\
 & Roberta & \textbf{61.88} &  \\ \cline{1-3}
\multirow{3}{*}{\begin{tabular}[c]{@{}c@{}}Question Agent Pairing\\  (Descriptions)\end{tabular}} & BM25 & 23.69 &  \\
 & USE & \textbf{43.59} &  \\
 & Roberta+STS & 36.67 &  \\ \cline{1-3}
\multirow{5}{*}{Response Selection} & BM25 & 59.94 &  \\
 & USE & 64.42 &  \\
 & USE QA & 71.66 &  \\
 & Roberta+STS & 56.82 &  \\
 & MARS & \textbf{83.55} &  \\ \cline{1-3}
\multirow{4}{*}{Individual Agents} & Alexa & 44.09 &  \\
 & Google & \textbf{48.06} &  \\
 & Houndify & 32.04 &  \\
 & Adasa & 3.45 &  \\ \hline
\end{tabular}
}
\caption{Performance breakdown of QA and QR approaches on our BBAI task on all 19 commercial agents we show that the MARS encoder is able to scale and leverage the capabilities of new agents added to the ensemble without diminishing performance compared to other approaches.}
\label{tab:results-19}
\vspace{-.6cm}
\end{table}

\section{Results and Discussion}

In this section we present and analyze the results of our experiments, detailing our insights and discussing the implications of each of our techniques. 

\paragraph{Evaluation task:} Similar to standard information retrieval evaluation measures, we denote accuracy as the metric \emph{precision@1} and use it to evaluate both our question agent and question response pairing approaches. For question agent pairing this metric denotes: Given a set of $N$ agents to the given query, whether the agent selected ultimately resolves the query successfully. For question response pairing it denotes: Given a set of $N$ responses to the given query, whether the top-scoring response resolves the query successfully. For this evaluation, we test on examples with at least one valid agent response.

\subsection{Question agent pairing} The results are summarized in tables \ref{tab:results-4} and \ref{tab:results-19}. We find that for the QA pairing Roberta yields the best result with an accuracy of 69\% in selecting the correct agent and 61.8\% when scaled to 19 agents. Similarly, we see that we can achieve fair performance in extreme data-scarce environments when using simple agent descriptions compared to that of query agent examples, with USE achieving 47.8\% accuracy. Using agent descriptions offers greater flexibility in facilitating the improvement of agents over time compared to query examples since it only requires an update to the agent description. However, it still falls short when compared to using a single agent like Google or Alexa. Also, while consistent in learning to recognize the domain a given agent may be performant in, QA approaches fall short in a few cases: 

\textbf{(1) Agent overlap} - This is when a given domains' coverage is split between various agents. e.g The model learns that both Alexa \& Google have proficiency in handling some weather queries but it remains unclear about which one is best suited for the current query at hand. 

\textbf{(2) Query variation} - While an agent's examples or descriptions may allude to proficiency in a given domain, it may still fail when asked certain query variations. e.g Figure \ref{fig:ofa} shows a case where Alexa is capable of handling weather queries but fails when a condition like humidity is asked for. Another example is when a similar question is asked in a different or more complex way. Both Houndify \& Alexa are known to be proficient at answering age-related questions but for questions like \textit{"How old I will be on September 28, 1995, if I was born on March 29, 1967?"}, Alexa is unable to answer as opposed to Houndify.

These cases are further highlighted when inspecting QA pairing performance at the domain level in table \ref{tab:domain}. We find that the QA approaches struggle with domains such as \emph{"travel suggestion"} and \emph{"Directions"} which are heavily split in coverage and more diverse in their variation.

\begin{table}[]
\centering
\resizebox{\columnwidth}{!}{%
\begin{tabular}{|lrrr|}
\hline
  \multicolumn{4}{|c|}{\textbf{Evaluation Performance per Domain (n=19)}} \\ \hline
\textbf{Domain} & \multicolumn{1}{l|}{\textbf{MARS (QR)}} & \multicolumn{1}{l|}{\textbf{USE (QA)}} & \multicolumn{1}{l|}{\textbf{Roberta (QA)}} \\ \hline
Weather & \multicolumn{1}{r|}{0.88} & \multicolumn{1}{r|}{0.45} & 0.67 \\ \hline
Directions & \multicolumn{1}{r|}{0.78} & \multicolumn{1}{r|}{0.29} & 0.44 \\ \hline
Auto & \multicolumn{1}{r|}{1.00} & \multicolumn{1}{r|}{0.79} & 0.82 \\ \hline
Restaurant Suggestion & \multicolumn{1}{r|}{0.79} & \multicolumn{1}{r|}{0.5} & 0.68 \\ \hline
Travel Suggestion & \multicolumn{1}{r|}{0.97} & \multicolumn{1}{r|}{0.33} & 0.57 \\ \hline
Time & \multicolumn{1}{r|}{0.81} & \multicolumn{1}{r|}{0.54} & 0.76 \\ \hline
Flight Info & \multicolumn{1}{r|}{0.83} & \multicolumn{1}{r|}{0.61} & 0.7 \\ \hline
Date & \multicolumn{1}{r|}{0.82} & \multicolumn{1}{r|}{0.47} & 0.56 \\ \hline
\end{tabular}
}
\caption{Further breakdown of the best-performing approaches per technique on a subset of 8 out of the 37 domains. We find that our MARS encoder generalizes well across the various agent domains.}
\label{tab:domain}
\vspace{-.5cm}
\end{table}

\subsection{Question response pairing}
In overall performance we find that our MARS encoder outperforms strong baselines, achieving 83.55\% accuracy on the BBAI task. We note that our MARS encoder outperforms the best single performing agent (Google Assistant) by 32\%. This shows the utility and power of OFA in not only alleviating the need for users to learn and adopt multiple agents but also validating that multiple agents working collectively can achieve significantly more than single agents working in isolation.

When inspecting the performance of MARS at the domain level we see in Table \ref{tab:domain} that it is able to maintain its high performance across the varying domains unlike the QA approaches. This advantage comes from the ability to select an agent at the response level allowing the system to catch cases in which an agent once deemed proficient fails or another agent improves.

\subsection{Agent pairing vs Response pairing}
We now describe the trade-offs between agent pairing and response pairing. Question response pairing greatly outperforms agent pairing in terms of accuracy, given that it is privy to the final responses from each of the agents. However, in practice, this comes with additional networking, compute, and latency costs from having to send the query to each of the agents and await their response. Given that the querying of agents is done in parallel, the latency cost is equal to that of the slowest agent. 
Question response pairing also better supports agent adaptation. With response pairing, a system can seamlessly add or remove an agent without diminishing the experience as shown by MARS in table \ref{tab:results-19}. In addition, as conversational agents are upgraded to offer a more diverse feature-set such as new domain support or improved responses, they can instantly be integrated into a response pairing approach.

\subsection{Scalability}
We evaluate our approaches on a suite of 19 commercially deployed agents spanning 37 broad domain categories. As shown in table \ref{tab:results-4} we examine performance when using the 4 largest agents in terms of domain support and popularity (Alexa, Google Assistant, Houndify and Ford Adasa) showing improvement upon single-agent use in both QA and QR approaches. When scaled up to 19 agents, MARS encoder improves even further by leveraging the new capabilities of the additional agents and is the only approach that does not decrease in performance as the number of agents and domains scale. This improvement is due to the input sensitive representations that the MARS encoder is able to learn by encoding both the question and response in a single transformer.

\textbf{Cross-encoding vs Bi-encoding}
For pairwise sentence scoring tasks such as response selection which compare question response pairs, it is important to be mindful of the trade-offs between cross encoder based models such as MARS in figure \ref{fig:system} (b) and bi-encoder models such as USE in \ref{fig:system} (a). Cross-encoders perform full self-attention over the pairwise input of the question and response, thus, producing an encoding representative of the combined input. This typically leads to much more performative models, especially in pairwise scoring tasks such as ours. However, given that this encoding isn't independent of the question for each question response pair, it is necessary to produce an encoding for each question label pair. Bi-encoders on the other hand perform self-attention over the question response pairs separately, map them to a dense vector space, and score them using an appropriate distance metric. With this separation, bi-encoders are able to index the question and compare these representations for each response resulting in faster prediction times when the numbers of candidate responses to a given question scales. Given the nature of our BBAI task which focuses on the scoring of responses to a singular question as opposed to a clustering task which requires an encoding for every pairwise combination across a set of sentences, cross-encoder based architectures remain a viable option even at the production scale for our use case.

\section{Related Work}

Ensemble approaches to solving complex tasks in the context of NLP are widely used \cite{deng2014ensemble, ARAQUE2017236}.
In dialogue systems, recent attempts at ensemble approaches and multi-agent architectures include \citet{alanav2} and \citet{Subramaniam:2018:CCM:3237383.3237472}.
AlanaV2 \cite{alanav2} demonstrated an ensemble architecture of multiple bots using a combination of rule-based machine learning systems built to support topic-based conversations across domains. It was built to be an open domain bot supporting topic-based conversations. Specifically, AlanaV2's architecture utilizes a variety of ontologies and NLU pipelines that draw information from a variety of web sources such as Reddit. However, its agent selection approach is guided by a simple priority bot list.
Subramaniam et al.~\cite{Subramaniam:2018:CCM:3237383.3237472} describe their conversational framework that employs an \textit{Orchestrator Bot} to understand the user query and direct them to a domain-specific bot that handles subsequent dialogue. In our work, we expand up the multi-agent goal by focusing on the integration of black-box conversational agents at scale.

\subsection{Response Selection}
This is the task of selecting the most appropriate response given context from a pool of candidates. It is a central component of information retrieval applications and has become a focal point in the evaluation of dialogue systems. \cite{sato-etal-2020-evaluating, henderson2019convert, DBLP:journals/corr/abs-2010-07785}. Prior work has shown strong performance on sentence pairing tasks through the use of sentence encoders and language model fine-tuning \cite{henderson2019convert, Humeau2020Poly-encoders:, reimers-2019-sentence-bert}. In our work, we explore the task of response selection using it as one of the bases for integrating black-box conversation agents.

\section{Conclusion}
The rapid proliferation of conversational agents calls for a unified approach to interacting with multiple CAs. The key challenge of building such an interface lie in that most commercial CAs are black-boxes with hidden internals. This paper introduces BBAI a new task of agent integration that focuses on unifying black-boxes CAs across varying domains. We explore two task techniques, question agent pairing and question response pairing and present One For All, a scalable system that unifies multiple black-box CAs with a centralized user interface. Using a combination of commercially available conversational agents, we evaluate a variety of approaches to multi-agent integration through One For All.
Our MARS encoder achieves 83.5\% accuracy on BBAI and outperforms the best single agent configuration by over 32\%. These results demonstrate the power of One For All which can leverage state-of-the-art NLU approaches to enable multiple agents to collectively achieve more than any single conversational agent in isolation eliminating the need for users to learn and adopt multiple agents. 

This work opens up a wide range of potential future work involving the design of systems geared towards facilitating more advanced multi-agent interaction. We foresee a system with even greater response selection performance as the NLP community continues to produce more state-of-the-art language models with even greater contextual knowledge of the world. Extensions of this work can include examining not only the integration of agents but the interoperability by facilitating the passing of shared conversation knowledge across agents especially in multi-turn conversational scenarios across multiple agents.

\section*{Acknowledgements}
We thank our anonymous reviewers for their feedback and suggestions. We also thank Yi-Chun Chen who assisted in designing the diagrams and figures shown in this work. This work was sponsored by Ford Motor Company.

\bibliography{anthology,custom}
\bibliographystyle{acl_natbib}




\end{document}